\documentclass[10pt]{article} 
\usepackage[preprint]{tmlr}

\usepackage{amsmath,amsfonts,bm}

\def\eqref#1{equation~\ref{#1}}

\def\1{\bm{1}}

\DeclareMathAlphabet{\mathsfit}{\encodingdefault}{\sfdefault}{m}{sl}
\SetMathAlphabet{\mathsfit}{bold}{\encodingdefault}{\sfdefault}{bx}{n}

\usepackage{hyperref}
\usepackage{url}
\usepackage[table]{xcolor}
\usepackage{xcolor}
\usepackage{colortbl}
\usepackage{booktabs}
\usepackage{float} 
\usepackage{algorithm}
\usepackage{algorithmicx}
\usepackage{listings}
\usepackage{algcompatible}
\definecolor{best}{HTML}{F8B2A2}    
\definecolor{second}{HTML}{F6DFD6}  
\usepackage{multirow}
\usepackage[table]{xcolor}
\usepackage{xcolor}
\usepackage{colortbl}
\usepackage{float} 
\usepackage{algorithm}
\usepackage{algorithmicx}
\usepackage{listings}
\usepackage{algcompatible}
\definecolor{best}{HTML}{F8B2A2}    
\definecolor{second}{HTML}{F6DFD6}  
\usepackage{multirow}
\usepackage{svg}

\title{Skill-RAG: Failure-State-Aware Retrieval Augmentation via
Hidden-State Probing and Skill Routing}

\author{
\name Kai Wei \email weikai@umich.edu \\
\addr University of Michigan
\AND
\name Raymond Li \email raymondl@cs.ubc.ca \\
\addr University of British Columbia
\AND
\name Xi Zhu \email xi.zhu@rutgers.edu \\
\addr Rutgers University
\AND
\name Zhaoqian Xue \email zhaoqian.xue@pennmedicine.upenn.edu \\
\addr University of Pennsylvania
\AND
\name Jiaojiao Han \email liuliujiujiu05@gmail.com \\
\addr New Jersey Institute of Technology
\AND
\name Jingcheng Niu \email jingcheng.niu@tu-darmstadt.de \\
\addr Independent Researcher
\AND
\name Fan Yang \email yangfan@wfu.edu \\
\addr Wake Forest University
}

\def\month{Jun}  
\def\year{2026} 

\begin{document}

\maketitle

\begin{abstract}
Retrieval-Augmented Generation (RAG) grounds large language models in external knowledge by querying document collections at inference time. While adaptive retrieval has improved efficiency, existing approaches treat post-retrieval failure as a retry signal rather than a diagnostic one, leaving the structural causes of query-evidence misalignment unaddressed. Most persistent retrieval failures stem not from missing evidence but from a mismatch between the query and the evidence space. We propose \textbf{Skill-RAG}, a failure-aware RAG framework that couples a lightweight hidden-state probe with a prompt-based skill router. The probe gates retrieval at two pipeline stages. On detecting a failure, the skill router diagnoses the cause and selects one of four corrective skills: query rewriting, question decomposition, evidence focusing, or an exit skill for truly irreducible cases. Across open-domain QA and complex reasoning benchmarks, Skill-RAG improves accuracy on hard cases that persist after multi-turn retrieval, with the largest gains on out-of-distribution datasets. Representation-space analyses show that the four skills occupy structured, separable regions of the failure state space, confirming that query-evidence misalignment is typed rather than monolithic. Our code is available at:{\color{blue}\url{https://github.com/weikai202/SkillRAG}.}
\end{abstract}

\section{Introduction}
Retrieval-augmented generation (RAG) improves factual reliability by grounding large language models (LLMs) in dynamically retrieved external knowledge \citep{lewis2020retrieval, gao2023retrieval, karpukhin2020dense, guu2020retrieval, guo2026deepsieve, jin2025disentangling}. Standard RAG retrieves once per query regardless of need. Adaptive retrieval methods break this uniformity by deciding whether, when, or how often retrieval should be invoked. For example, some methods classify query complexity to determine whether retrieval is needed \citep{jeong2024adaptiverag, mallen2023not}, while others train models to emit explicit retrieval-trigger tokens during decoding \citep{asai2024selfrag}. Iterative methods instead perform multiple retrieval rounds per query, issuing follow-up queries to progressively refine the context \citep{jiang2023active}. Another related line inspects the model's internal state during generation, using attention entropy or probes over hidden representations to detect uncertainty and trigger retrieval only when the current context is insufficient \citep{su-etal-2024-dragin, baek2025probing}.

However, existing methods treat retrieval control as a coarse-grained decision (whether to retrieve and how many times), overlooking the structural causes of retrieval failure and the corrective strategies they require. Queries requiring multi-step reasoning often fail because each retrieval step depends on synthesizing evidence from previous steps, a dependency that repeated retrieval alone cannot satisfy \citep{trivedi2023interleaving, tang2024multihop, press2023measuring, yao2023react}. Noisy or irrelevant retrieved passages can further degrade performance by distracting the model from the target answer \citep{shi2023large}. These failures stem not from missing evidence but from structural alignment gaps: queries poorly specified for the evidence space, causing retrieved documents to be topically related but inferentially insufficient \citep{shi2023large, liu2024lost}. These failures exhibit structured patterns in the model's internal representations \citep{gao2025probing, jin2025exploring, jin2026farther, han2026sage} (e.g., overly broad queries requiring evidence focusing \citep{yan2024corrective}, entangled premises requiring decomposition \citep{press2023measuring, khattab2022demonstrate}, and divergent surface forms requiring rewriting \citep{ma2023query, gao2023precise}). Section~\ref{subsec:failure_analysis} demonstrates this geometric structure empirically. We coin the term \textit{failure states} for these latent signals: latent representations extracted from the hidden layers indicate when retrieval has stalled. These states enable a skill router to select targeted retrieval actions \citep{asai2023task}.

Therefore, we propose \textbf{Skill-RAG}, a failure-aware RAG framework in which a lightweight hidden-state probe detects when retrieval has stalled and gates entry into skill routing. Upon detecting a failure state, a prompt-based skill router diagnoses the underlying cause and selects one of four retrieval skills. These are query rewriting \citep{ma2023query}, question decomposition \citep{press2023measuring}, evidence focusing \citep{yan2024corrective}, and an exit skill that identifies genuinely unresolvable cases and terminates retrieval gracefully \citep{feng2024don, kadavath2022language}. Unlike prior work that optimizes retrieval triggering or depth \citep{shao2023enhancing, asai2024selfrag, shao2023enhancing}, Skill-RAG reframes post-retrieval recovery as a conditional skill selection problem. This provides fine-grained, failure-conditioned control over how LLMs access external knowledge. Three contributions follow: \textbf{(1)} We propose the first probe-and-route pipeline for post-retrieval failure recovery, combining hidden-state probe gating with a prompt-based skill router without additional LLM calls; \textbf{(2)} We introduce a four-skill retrieval vocabulary grounded in observed failure patterns and validate it across different distribution datasets, where the exit skill saves 1.42 retrieval rounds on average at a cost of 0.6 accuracy points; \textbf{(3)} We evaluate Skill-RAG on three backbone LLMs (Gemma2-9B, Qwen3-8B, Llama3-8B), achieving state-of-the-art performance on in-domain benchmarks and outperforming the probe-only baseline \citep{baek2025probing} by up to 13.6 ACC points on OOD datasets.

\section{Preliminaries}

In this section, we introduce the building blocks of our retrieval gating mechanism. We first review evidence that LLM hidden states encode reliable signals about the model's knowledge state (\S\ref{sub:hidden_states_as_knowledge_signials}), then describe the probe that reads these signals (\S2.2), and finally specify how its predictions gate retrieval at inference time (\S2.3).

\begin{figure*}[t]
  \includegraphics[width=1\linewidth]{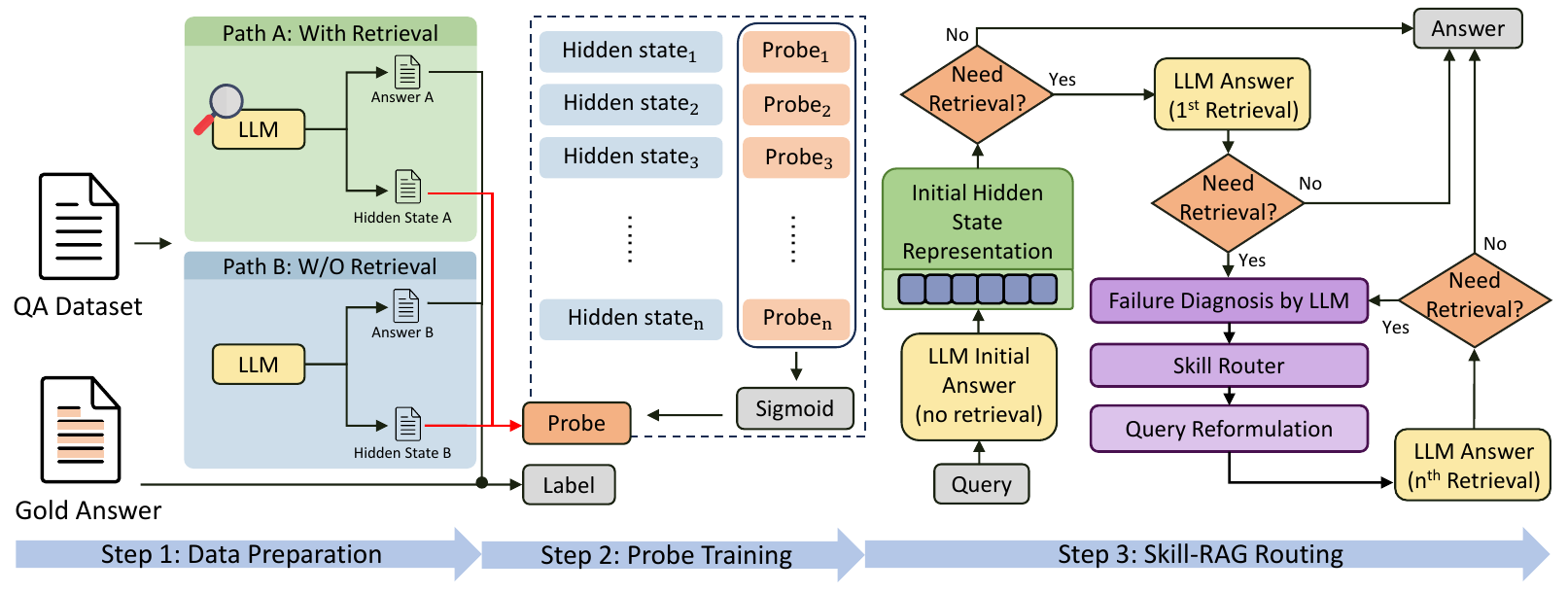}
  \caption{Skill-RAG Pipeline. \textnormal{\textbf{Step 1:} For each in-domain training example, we generate answers under two conditions (no retrieval and single-step retrieval) and extract the resulting hidden states. \textbf{Step 2:}  Binary correctness labels, derived by comparing generated answers against gold answers, train the probe to predict whether retrieval is needed. \textbf{Step 3:}  At each model output, the probe evaluates the current hidden states to determine whether retrieval is needed. On each retrieval failure, the skill router diagnoses the misalignment cause and selects a corrective retrieval skill. This loop continues until the probe confirms sufficiency or the retrieval budget is exhausted.}}
  
  \label{fig:1}
\end{figure*}

\subsection{Hidden States as Knowledge Signals}
\label{sub:hidden_states_as_knowledge_signials}
The intermediate representations of LLMs encode information about the model's knowledge state that is not only reflected in the final output distribution. Probing studies show that linear classifiers trained on hidden-layer activations reliably detect whether a statement is truthful before decoding \citep{azaria2023internal, marks2023geometry, jin2026farther}. These truth-encoding directions transfer across domains. Unsupervised methods such as Contrast-Consistent Search recover latent knowledge directly from activation geometry without labeled supervision \citep{burns2022discovering}. \citet{li2023inference} demonstrate that shifting activations along learned truthful directions at inference time (without retraining) substantially improves factual accuracy. This confirms that the relevant signal is causally upstream of the output distribution. Attention-weight ratios over context versus newly generated tokens provide a lightweight, reliable signal for contextual hallucination \citep{chuang2024lookback}. Contrasting logits from earlier and later layers reduces factual errors in generation \citep{chuang2023dola}. These results establish a single principle: \textit{what the model knows can be read from its internal states before generation completes}. We adopt this perspective to gate retrieval in RAG systems, training a lightweight hidden-state probe to detect when retrieval has stalled rather than relying on output confidence.

\subsection{Probe Architecture}
Following \citet{baek2025probing}, we train a lightweight probe on the LLM's hidden states to predict whether the model can correctly answer a query. Given a generated sequence of reasoning and answer tokens ${t_1, \ldots, t_n}$, we extract hidden states from the final two-thirds of the model's layers. We sample every other layer to balance coverage and efficiency. Let $\mathbf{H}_l \in \mathbb{R}^{n \times d}$ denote the hidden states at layer $l$, where $d$ is the hidden dimension. We obtain a layer-level representation via mean pooling:
\begin{equation}
    \mathbf{h}_l = \frac{1}{n}\sum_{i=1}^{n} \mathbf{H}_l^{(i)}, \quad l \in \mathcal{S},
\end{equation}
where $\mathcal{S} = \{l : l \geq \lfloor L/3 \rfloor,\ l\ \text{even}\}$ is the set of selected layers and $L$ is the total number of layers. We pass each layer representation through a feed-forward probe with a single hidden layer:
\begin{equation}
    p_l = \sigma\left(\mathbf{W}_2 \cdot \text{ReLU}(\mathbf{W}_1 \mathbf{h}_l + \mathbf{b}_1) + b_2\right),
\end{equation}
where $\sigma$ denotes the sigmoid function. Finally, we aggregated per-layer probabilities by averaging across selected layers.
\begin{equation}
    \hat{p} = \frac{1}{|\mathcal{S}|}\sum_{l \in \mathcal{S}} p_l.
\end{equation}

\subsection{Gating Decision}
At inference time, we compare $\hat{p}$ against a fixed threshold $\tau$. If $\hat{p} < \tau$, the model lacks sufficient knowledge for correct generation and retrieval proceeds; otherwise, the system returns the answer directly. We train the probe on hidden states collected under two retrieval conditions (no retrieval and single-step retrieval), assigning binary labels by comparing generated answers against gold answers. This mechanism requires no additional LLM calls and adds negligible inference overhead, serving as the gating foundation for the skill routing framework described in \S\ref{sec:method}.

\section{Method}
\label{sec:method}
Figure~\ref{fig:1} illustrates our proposed Skill-RAG pipeline. Given a query, a hidden-state probe first assesses whether the model's parametric knowledge suffices to answer without retrieval; if so, the system returns the answer directly. Otherwise, retrieval proceeds and the probe re-evaluates the augmented generation. If the retrieved evidence suffices, the pipeline yields the answer. On failure, the skill router receives the reasoning trace, answer, and retrieved evidence and diagnoses the misalignment cause. It then selects one of four retrieval skills to reformulate the query or refocus the evidence. The revised query triggers a new retrieval round, and the probe gates the next iteration. This loop continues until the probe judges the model's state sufficient or the retrieval budget is exhausted.

\subsection{Probe Training}
To train the probe, we run two retrieval strategies on the training splits of HotpotQA \citep{yang-etal-2018-hotpotqa}, NQ \citep{kwiatkowski-etal-2019-natural}, and TriviaQA \citep{joshi2017triviaqa}: no retrieval and single-step retrieval. In both cases, we prompt the model to produce a chain-of-thought reasoning trace followed by a final answer. For each training example, we extract hidden states at reasoning and answer tokens from the final two-thirds of the model's layers. Comparing the generated answer against the gold answer assigns a binary label, yielding a dataset of hidden-state representations with correctness labels. The probe is a feed-forward network with a single hidden layer and a binary classification head. To capture information across layers, we train one probe per layer; at inference time, we average their predicted probabilities into a single gating signal reflecting answer readiness across the model's depth.

\subsection{Skill Router and Termination}
When the probe detects a failure state, it invokes the prompt-based skill router. Given the original question, the model's failed reasoning and answer, and the retrieved evidence, the router diagnoses the cause of misalignment and selects one of four retrieval skills. \textbf{Query rewriting} targets cases where the query's surface form diverges from corpus indexing conventions, producing a reformulated query better aligned with retrievable evidence \citep{ma2023query}. \textbf{Question decomposition} addresses multi-hop queries with entangled premises, generating a sequence of sub-queries that isolate each reasoning step before issuing a final retrieval query \citep{press2023measuring}. \textbf{Evidence focusing} handles semantically broad queries by extracting missing evidence slots from the current context and issuing a targeted query for the identified gap \citep{yan2024corrective}. \textbf{Exit} identifies cases where misalignment is irreducible, due to missing knowledge or model capacity limits, and terminates retrieval to avoid unnecessary inference overhead.

Following skill execution, the retriever receives the reformulated query, and the model generates a new answer conditioned on the updated evidence. The probe then gates the next iteration. This loop continues until the skill router selects exit, the probe judges the model's state sufficient, or the retrieval budget is exhausted.

\section{Experiments}

Now, we empirically evaluate Skill-RAG and analyze its gains. After describing the experimental setup (§4.1), we report main results on five QA benchmarks (§4.2), examine the geometric structure of failure cases that motivates our skill vocabulary (§4.3), and quantify the efficiency benefits of the Exit skill (§4.4).

\subsection{Experimental Setup}
We evaluate on five open-domain QA benchmarks spanning single and multi-hop reasoning. Three datasets (NQ \citep{kwiatkowski-etal-2019-natural}, TriviaQA \citep{joshi2017triviaqa}, and HotpotQA \citep{yang-etal-2018-hotpotqa}) serve as in-domain benchmarks, from which we sample 3,000 examples for probe training and 500 for development. We hold out two multi-hop datasets: MuSiQue \citep{trivedi2022musique} and 2WikiMultiHopQA \citep{ho2020constructing}, as out-of-distribution test sets, evaluating on 500 examples each. All methods use BM25 \citep{robertson2009probabilistic} as the retriever.

We compare Skill-RAG against six baselines: \textbf{No Retrieval}, which generates answers from parametric knowledge alone; \textbf{Single-step RAG}, which performs one round of retrieval before generation; \textbf{FLARE} \citep{jiang2023active}, which triggers retrieval based on token-level generation uncertainty; \textbf{DRAGIN} \citep{su-etal-2024-dragin}, which determines retrieval timing via attention-based relevance signals; \textbf{Adaptive-RAG} \citep{jeong2024adaptiverag}, which routes queries to retrieval strategies of varying complexity via a trained classifier; and \textbf{Probing-RAG} \citep{baek2025probing}, which gates retrieval decisions using hidden-state probing. All experiments are conducted on three backbone models: Gemma2-9B, Qwen3-8B, and Llama3-8B, under 4-shot prompting. We report both Exact Match and Accuracy for all evaluations.

\begin{table}[t]
\centering
\label{tab:main_results}
\setlength{\tabcolsep}{3.8pt}
\renewcommand{\arraystretch}{0.95}
\scriptsize
\resizebox{0.8\textwidth}{!}{
\begin{tabular}{lcccccccccc}
\toprule
\multirow{3}{*}{\textbf{Method}}
  & \multicolumn{6}{c}{\textbf{In-Domain}}
  & \multicolumn{4}{c}{\textbf{Out-of-Domain}} \\
\cmidrule(lr){2-7}\cmidrule(lr){8-11}
  & \multicolumn{2}{c}{HotpotQA}
  & \multicolumn{2}{c}{NQ}
  & \multicolumn{2}{c}{TriviaQA}
  & \multicolumn{2}{c}{MuSiQue}
  & \multicolumn{2}{c}{2Wiki} \\
  & EM & AC & EM & AC & EM & AC & EM & AC & EM & AC \\
\midrule

\multicolumn{11}{c}{\textit{Llama3-8B}} \\
\cmidrule(lr){1-11}
No Retrieval     & 22.3 & 34.6 & 29.8 & 49.4 & 50.1 & 58.7 & 7.2  & 11.6 & 26.4 & 41.8 \\
Single-step      & \underline{31.8} & \textbf{48.6} & 30.3 & 46.8 & 44.8 & 49.2 & 8.9  & 13.9 & \underline{28.9} & 46.3 \\
\cmidrule(lr){1-11}
FLARE            & 27.4 & 32.8 & 28.3 & 44.6 & 45.2 & 53.8 & 6.4  & 10.1 & 23.2 & 40.3 \\
DRAGIN           & \textbf{31.9} & 38.4 & \underline{31.6} & 50.2 & 51.8 & 57.6 & 8.9  & 13.4 & 27.8 & 46.7 \\
Adaptive-RAG     & 27.8 & 34.2 & 28.6 & 47.3 & 46.4 & 54.9 & 7.3  & 11.2 & 24.8 & 42.6 \\
Probing-RAG      & 21.4 & 43.2 & 30.1 & \underline{52.6} & \underline{55.0} & \underline{60.4} & \underline{9.2} & \underline{15.1} & 25.3 & \underline{47.8} \\
\textbf{Skill-RAG} & 23.1 & \underline{44.8} & \textbf{31.8} & \textbf{53.4} & \textbf{55.8} & \textbf{62.4} & \textbf{11.3} & \textbf{19.2} & \textbf{29.4} & \textbf{50.6} \\

\midrule
\multicolumn{11}{c}{\textit{Qwen3-8B}} \\
\cmidrule(lr){1-11}
No Retrieval     & 28.3 & 37.4 & 30.6 & 47.8 & 50.2 & 58.3 & 8.4  & 12.6 & 24.7 & 40.2 \\
Single-step      & 31.5 & 51.3 & 32.4 & 57.8 & 47.6 & 62.4 & \underline{11.8} & 18.3 & 27.3 & 44.6 \\
\cmidrule(lr){1-11}
FLARE            & 26.8 & 35.2 & 29.4 & 46.3 & 47.6 & 56.4 & 7.1  & 11.4 & 22.8 & 39.6 \\
DRAGIN           & 32.4 & 40.1 & 33.8 & 52.7 & 51.4 & 60.8 & 10.5 & 14.7 & 26.4 & 43.8 \\
Adaptive-RAG     & 28.6 & 36.5 & 30.2 & 49.1 & 48.3 & 57.9 & 8.6  & 12.3 & 24.1 & 41.5 \\
Probing-RAG      & \underline{33.8} & \underline{54.6} & \underline{34.2} & \underline{59.3} & \underline{52.4} & \underline{65.7} & 10.3 & \underline{18.9} & \underline{27.8} & \underline{45.3} \\
\textbf{Skill-RAG} & \textbf{34.5} & \textbf{55.8} & \textbf{35.1} & \textbf{60.1} & \textbf{53.2} & \textbf{65.9} & \textbf{15.6} & \textbf{25.3} & \textbf{34.2} & \textbf{54.8} \\

\midrule
\multicolumn{11}{c}{\textit{Gemma2-9B}} \\
\cmidrule(lr){1-11}
No Retrieval     & 24.6 & 37.1 & 27.3 & 42.7 & 52.4 & \underline{65.7} & 6.1  & 9.8  & \underline{29.3} & \underline{44.5} \\
Single-step      & 27.8 & 42.6 & 26.1 & 40.7 & 46.5 & 59.3 & 6.4  & \underline{15.2} & 26.5 & 43.8 \\
\cmidrule(lr){1-11}
FLARE            & \underline{29.5} & 33.8 & 29.7 & 36.4 & 44.8 & 57.2 & 5.3  & 8.4  & 24.1 & 36.8 \\
DRAGIN           & \textbf{35.6} & 37.5 & \underline{34.2} & 38.9 & \underline{55.3} & 63.4 & 7.4  & 11.2 & \textbf{30.8} & 42.3 \\
Adaptive-RAG     & 29.1 & 32.3 & 29.4 & 34.5 & 46.2 & 58.6 & 5.8  & 9.1  & 25.6 & 38.4 \\
Probing-RAG      & 22.8 & \underline{44.7} & 34.1 & \underline{48.5} & \textbf{59.3} & \textbf{65.9} & \underline{7.6} & 13.9 & 26.6 & 38.9 \\
\textbf{Skill-RAG} & 24.2 & \textbf{46.1} & \textbf{34.3} & \textbf{49.7} & \textbf{59.3} & \textbf{65.9} & \textbf{7.8} & \textbf{20.0} & 28.9 & \textbf{52.5} \\
\bottomrule
\end{tabular}
}
\caption{Main results on in-domain and out-of-distribution benchmarks.
\textnormal{\textbf{Bold} indicates the best result and \underline{underline} indicates the second best. 
Skill-RAG preserves in-domain performance and generalizes substantially to OOD benchmarks, 
where skill routing outperforms probe gating alone.}}
\end{table}

\subsection{Main Results}
Table~\ref{tab:main_results} reports results on three models in five benchmarks. Skill-RAG matches or surpasses Probing-RAG on both EM and ACC across in-domain benchmarks and achieves the largest gains on OOD benchmarks. On Llama3-8B, Skill-RAG trails only on HotpotQA: DRAGIN reaches EM 31.9 and the single-step method reaches ACC 48.6. Qwen3-8B shows the clearest gains: Skill-RAG outperforms Probing-RAG by \textbf{6.4} ACC and \textbf{5.3} EM on MuSiQue and by \textbf{9.5} ACC and \textbf{6.4} EM on 2WikiMultiHopQA, achieving SOTA across all five benchmarks. On Gemma2-9B, Skill-RAG gains \textbf{6.1} ACC points over Probing-RAG on MuSiQue and \textbf{13.6} on 2WikiMultiHopQA. DRAGIN achieves higher EM on HotpotQA and 2WikiMultiHopQA, but Skill-RAG leads on ACC and in OOD settings. This advantage suggests that skill routing is most effective when failures require targeted alignment correction rather than repeated retrieval.

\begin{figure}[t]
  \centering
  \includegraphics[width=0.6\textwidth]{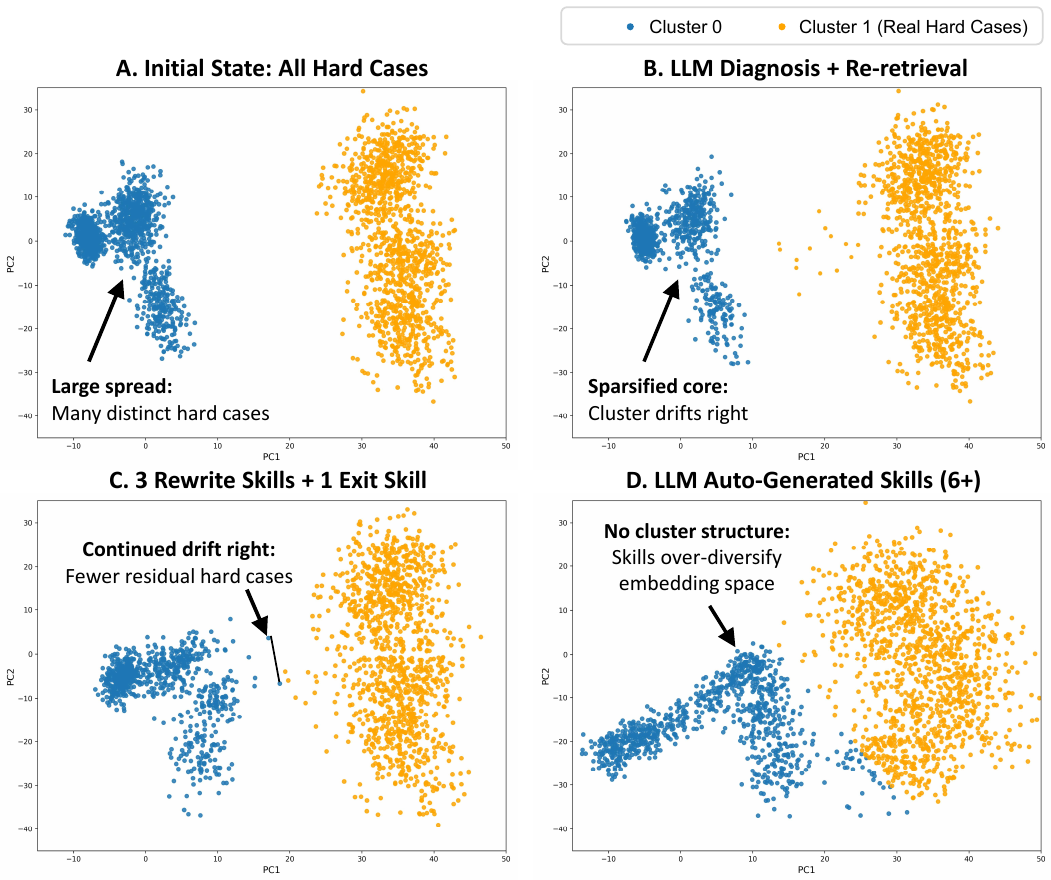}
  \caption{Hidden-State Embeddings Visualization. 
  \textnormal{The left cluster shifts rightward as skill count increases but collapses beyond a threshold.}}
  \label{fig:cluster}
\end{figure}

\subsection{Failure Representation Space Analysis}\label{subsec:failure_analysis}  

Figure~\ref{fig:cluster} visualizes t-SNE projections of hidden-state embeddings for cases that remain incorrect after three rounds of standard retrieval, across four experimental conditions. In the initial state (A), two geometrically separable clusters emerge. Applying skills progressively contracts the left cluster: re-retrieval alone (B) shifts some alignment-fixable cases rightward; the full four-skill vocabulary (C) shrinks the cluster to a small residual. Expanding beyond six auto-generated skills (D) dissolves the cluster entirely, merging both populations into an undifferentiated distribution. This collapse shows that over-diversification disrupts the representational geometry needed for targeted routing. The four-skill vocabulary is principled: it reflects the intrinsic structure of the failure space.

\begin{figure}[t]
    \centering
    \includegraphics[width=0.8\linewidth]{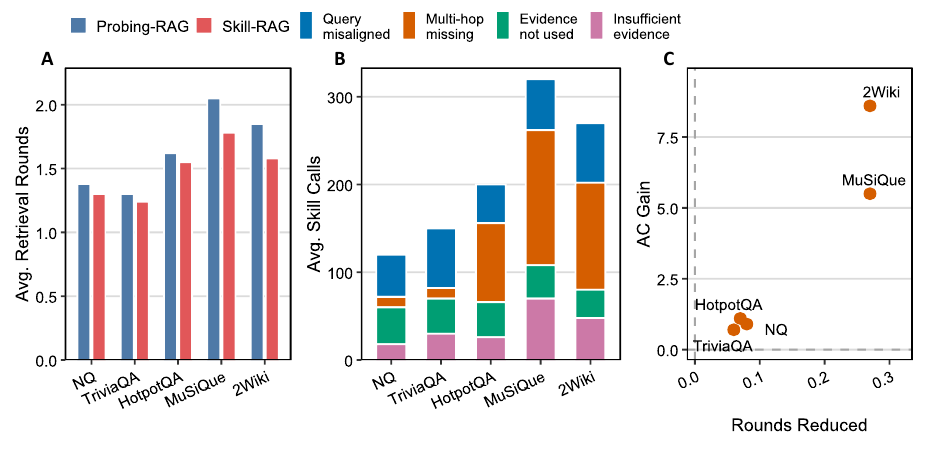}
    \caption{Retrieval efficiency and skill usage across datasets. 
    \textnormal{\textbf{(A)} Skill-RAG reduces average retrieval rounds on every dataset relative to the probe-only baseline. 
    \textbf{(B)} Skill-call distribution per dataset; the multi-hop skill's share is higher on OOD than in-domain datasets. 
    \textbf{(C)} Joint scatter of accuracy gain versus retrieval reduction per dataset; OOD datasets occupy the upper-right quadrant, gaining accuracy while cutting rounds.}}
    \label{fig3}
\end{figure}

\subsection{Retrieval Efficiency and Skill Routing}
\label{subsec:efficiency_analysis}  

Skill-RAG reduces retrieval rounds on every dataset (Figure~\ref{fig3}A, averaged across three backbone models). In-domain savings are modest (NQ: $1.38\to1.30$; TriviaQA: $1.30\to1.24$; HotpotQA: $1.62\to1.55$), while OOD sets save 0.27 rounds each (MuSiQue: $2.05\to1.78$; 2Wiki: $1.85\to1.58$), consistent with the higher exit-skill rate on OOD datasets. The skill distribution in Figure~\ref{fig3}B tracks each dataset's query structure. On NQ, query rewrite (40\%) and evidence grounding (35\%) dominate; TriviaQA raises the rewrite share to 45\%, and HotpotQA shifts to question decomposition (45\%). OOD datasets show the same shift: decomposition peaks at 48\% on MuSiQue and 45\% on 2Wiki. On MuSiQue, the exit skill also reaches its highest share across all datasets at 22\%, indicating that many queries resist resolution through further retrieval. Figure~\ref{fig3}C plots AC Gain ($\text{AC(Skill-RAG)} - \text{AC(Probing-RAG)}$) against Rounds Reduced ($\overline{\text{Rounds}}_{\text{Probing}} - \overline{\text{Rounds}}_{\text{Skill-RAG}}$) per dataset. In-domain datasets cluster near the origin ($<$1.1 AC points, $<$0.08 rounds saved); both OOD datasets occupy the upper-right: MuSiQue gains 5.5 points while saving 0.27 rounds, and 2Wiki gains 8.6 at the same saving. Benefit concentrates on OOD queries, where structural misalignment rather than evidence shortage drives failure.

We isolate all queries on which Probing-RAG exhausts its maximum retrieval budget and still answers incorrectly, then run these hard cases through Skill-RAG. Table~\ref{tab:early_exit_analysis} reports exit-skill selection rates, accuracy, and retrieval rounds with and without the exit skill. The exit-skill rate varies with model capacity and dataset difficulty: Qwen3-8b selects it most often (23.1\%), followed by Gemma2-9b (21.9\%) and LLaMA3-8b (19.8\%). OOD datasets show substantially higher rates than in-domain ones, ranging from 13\% on NQ to 28\% on MuSiQue. Across all conditions, early stopping costs 0.5--0.6 accuracy points relative to Skill-RAG without the exit skill, while saving 1.38--1.42 retrieval rounds per stopped query. The skill router thus plays two complementary roles on hard cases: corrective skills fix misalignment failures that further retrieval can resolve, while the exit skill terminates those that it cannot.

\begin{table}[t]
\centering
\setlength{\tabcolsep}{2.4pt}
\renewcommand{\arraystretch}{0.78}
\tiny
\resizebox{0.67\columnwidth}{!}{
\begin{tabular}{lcccccc}
\toprule
\textbf{Group} 
& \textbf{\#Cases} 
& \textbf{Exit} 
& \textbf{AC$_{\text{w/}}$} 
& \textbf{AC$_{\text{w/o}}$} 
& $\Delta$\textbf{AC} 
& \textbf{Saved} \\
\midrule

\multicolumn{7}{c}{\textit{By Dataset}} \\
\cmidrule(lr){1-7}
NQ        & 400  & 13.0\% & 8.8  & 9.3  & -0.5 & 1.35 \\
TriviaQA  & 300  & 17.0\% & 7.8  & 8.3  & -0.5 & 1.35 \\
HotpotQA  & 520  & 17.0\% & 10.0 & 10.5 & -0.5 & 1.37 \\
MuSiQue   & 815  & 28.0\% & 14.3 & 15.0 & -0.7 & 1.47 \\
2Wiki     & 625  & 23.0\% & 17.3 & 18.0 & -0.7 & 1.47 \\
\cmidrule(lr){1-7}
\textbf{All} 
& \textbf{2660} 
& \textbf{21.4\%} 
& \textbf{12.5} 
& \textbf{13.1} 
& \textbf{-0.6} 
& \textbf{1.42} \\

\midrule
\multicolumn{7}{c}{\textit{By Model}} \\
\cmidrule(lr){1-7}
Llama3-8B  & 945 & 19.8\% & 9.8  & 10.3 & -0.5 & 1.34 \\
Qwen3-8B   & 825 & 23.1\% & 14.6 & 15.2 & -0.6 & 1.39 \\
Gemma2-9B  & 890 & 21.9\% & 11.0 & 11.4 & -0.4 & 1.40 \\
\cmidrule(lr){1-7}
\textbf{All} 
& \textbf{2660} 
& \textbf{21.5\%} 
& \textbf{11.7} 
& \textbf{12.2} 
& \textbf{-0.5} 
& \textbf{1.38} \\
\bottomrule
\end{tabular}}
\vspace{0.3em}
\caption{
Early-exit analysis on hard cases. 
\textnormal{\textit{Exit} denotes the proportion of cases stopped before additional retrieval, and \textit{Saved} denotes the average number of retrieval rounds saved per exited case. \textsuperscript{*}All reported aggregate results are computed as weighted averages.}
}
\label{tab:early_exit_analysis}
\end{table}

\section{Conclusion}
We presented Skill-RAG, a failure-aware RAG framework that addresses a largely overlooked failure mode in adaptive retrieval: the structural query-evidence alignment gap that persists after initial retrieval. By coupling a lightweight hidden-state probe with a prompt-based skill router, Skill-RAG moves beyond binary retrieval triggering to provide targeted, failure-conditioned alignment correction. Experiments show OOD gains of up to 13.6 ACC points and consistent improvements on hard in-domain cases, confirming the generalization advantage of structured skill routing over probe gating alone. Representation-space analyses confirm that the proposed skill vocabulary reflects the intrinsic geometry of the failure state space. The geometric grounding suggests that interpretable, fine-grained retrieval control is tractable. Furthermore, exit effectively trades token cost against quality loss, but its benefit depends on data distribution and does not generalize to all settings. Future work will explore more adaptive exit calibration and extend the skill vocabulary to broader retrieval failures, including ambiguous questions, conflicting evidence, and domain-specific evidence formats, etc.

\appendix
\section{Appendix}
\subsection{Skill Prompts}
\label{appendix:prompts}

We provide the full prompt templates used by the skill router and each retrieval skill in Skill-RAG.

\vspace{1em}
\textbf{Diagnosis Prompt}
\label{appendix:diagnosis}

\begin{lstlisting}[basicstyle=\ttfamily\small, breaklines=true, frame=single]
You are analyzing why an answer is wrong. Read the question and the model's wrong reasoning+answer, then output one short error diagnosis sentence.

Question:
{question}

Wrong reasoning+answer:
{reasoning_answer}

Diagnosis:
\end{lstlisting}

\vspace{1em}
\textbf{Skill Router Prompt}
\label{appendix:router}

\begin{lstlisting}[basicstyle=\ttfamily\small, breaklines=true, frame=single]
You are a routing classifier.
Choose exactly one skill name from:
1) query_misaligned
2) multi_hop_missing
3) evidence_not_used
4) insufficient_evidence

Question:
{question}

Wrong reasoning+answer:
{reasoning_answer}

Diagnosis:
{diagnosis}

Only output one skill name.
If current evidences are clearly insufficient to answer correctly, output: 
insufficient_evidence
\end{lstlisting}

\vspace{1em}
\textbf{Skill 1: Query Rewriting}
\label{appendix:query_rewrite}

\begin{lstlisting}[basicstyle=\ttfamily\small, breaklines=true, frame=single]
You are a query rewrite expert for open-domain QA.
Task: rewrite ONE better search query based on failure context.

Example 1
Question: Which city hosts the Louvre Museum?
Failed reasoning+answer: I think Louvre is in Italy. 
Answer: Rome
Current evidences: passage 1: Louvre is a museum in Paris, France.
Search Query: city where Louvre Museum is located

Example 2
Question: Who wrote Pride and Prejudice?
Failed reasoning+answer: It might be Charles Dickens. 
Answer: Charles Dickens
Current evidences: passage 1: Pride and Prejudice is a novel by Jane Austen.
Search Query: author of Pride and Prejudice Jane Austen

Now solve:
Question: {question}
Failed reasoning+answer: {reasoning_answer}
Current evidences:
{evidences}

Output format:
Search Query: <one improved query>
\end{lstlisting}

\vspace{1em}
\textbf{Skill 2: Question Decomposition}
\label{appendix:decomposition}

\begin{lstlisting}[basicstyle=\ttfamily\small, breaklines=true, frame=single]
You are a decomposition expert for multi-hop retrieval.
Task: produce 2-3 sub-queries, then one final retrieval query.

Example 1
Question: Roger O. Egeberg served under which president, and that president served during what years?
Failed reasoning+answer: He served in health affairs. 
Answer: unknown years 
Current evidences: passage 1: Egeberg served during Nixon administration.
Sub-query 1: Which U.S. president is associated with the Nixon administration?
Sub-query 2: What are the presidency years of Richard Nixon?
Final Search Query: Richard Nixon presidency years 1969 1974

Example 2
Question: Which writer was from England, Henry Roth or Robert Erskine Childers? 
Failed reasoning+answer: Henry Roth is European. 
Answer: Henry Roth
Current evidences: passage 1: Henry Roth was an American novelist.
Sub-query 1: What is Henry Roth's nationality?
Sub-query 2: Where was Robert Erskine Childers born?
Final Search Query: Robert Erskine Childers born England nationality

Now solve:
Question: {question}
Failed reasoning+answer: {reasoning_answer}
Current evidences:
{evidences}

Output format:
Sub-query 1: ...
Sub-query 2: ...
Sub-query 3: ...
Final Search Query: <one combined query>
\end{lstlisting}

\vspace{1em}
\textbf{Skill 3: Evidence Focusing}
\label{appendix:evidence_focusing}

\begin{lstlisting}[basicstyle=\ttfamily\small, breaklines=true, frame=single]
You are an evidence-grounded retrieval planner.
Task: extract missing evidence slots, then output one grounded search query.

Example 1
Question: What years did the president in Egeberg's administration serve?
Failed reasoning+answer: Egeberg served under Nixon. 
Answer: 1970 to 1978
Current evidences: passage 1: Egeberg served during Nixon administration.
Missing Slot 1: Exact start and end years of Nixon presidency
Missing Slot 2: Reliable source confirming those years
Search Query: exact years Richard Nixon served as U.S. president

Example 2
Question: Are Giuseppe Verdi and Ambroise Thomas both opera composers?
Failed reasoning+answer: Verdi was a writer. 
Answer: No
Current evidences: passage 1: Ambroise Thomas is a French opera composer.
Missing Slot 1: Verdi's profession
Missing Slot 2: Confirmation both are opera composers
Search Query: Giuseppe Verdi profession opera composer evidence

Now solve:
Question: {question}
Failed reasoning+answer: {reasoning_answer}
Current evidences:
{evidences}

Output format:
Missing Slot 1: ...
Missing Slot 2: ...
Search Query: <one grounded query>
\end{lstlisting}

\vspace{1em}
\textbf{Skill 4: Insufficient Evidence Early Stop}
\label{appendix:insufficient}

\begin{lstlisting}[basicstyle=\ttfamily\small, breaklines=true, frame=single]
EXIT_SKILL = "insufficient_evidence"

selected_skill, diagnosis = choose_skill(question, current_reasoning_answer)

if selected_skill == EXIT_SKILL:
    log["selected_skill"] = EXIT_SKILL
    log["stopped_by_exit_skill"] = True
    log["exit_action"] = "stop_retrieval_use_current_answer"
    final_answer = current_reasoning_answer
    stop_retrieval = True
else:
    search_query = generate_search_query(
        selected_skill,
        question,
        current_reasoning_answer,
        current_evidences,
    )
    new_evidences = retrieve(search_query)
    final_answer = regenerate_answer(question, new_evidences)
\end{lstlisting}

\subsection{Case Study}

Table~\ref{tab:case_study} presents a concrete example that contrasts the workflows of Probing-RAG and Skill-RAG, highlighting how Skill-RAG applies targeted skill routing after detecting a retrieval failure.

\begin{table*}
\centering
\setlength{\tabcolsep}{3pt}
\renewcommand{\arraystretch}{1.3}
\begin{tabular}{p{0.10\linewidth} p{0.40\linewidth} p{0.40\linewidth}}
\toprule
& \textbf{ProbingRAG} & \textbf{Skill-RAG (Ours)} \\
\midrule
\multicolumn{3}{l}{%
  \textbf{Q:} \textit{What disease did the dancing plague of 1518 cause victims to suffer from?}
  \quad \textbf{Gold:} exhaustion} \\
\midrule
Initial
& \multicolumn{2}{p{0.82\linewidth}}{%
  Query: \textit{``dancing plague 1518 disease''}
  $\rightarrow$ No sufficient knowledge.
  Answer: \textit{``Unknown''} \textcolor{red}{ACC=0}} \\
\midrule
Round 1
& \multicolumn{2}{p{0.82\linewidth}}{%
  Query: \textit{``dancing plague 1518 disease''}
  \newline Evidence (shared): historical descriptions; no physical consequences mentioned.
  \newline Answer: \textit{``Dancing mania''} \textcolor{red}{ACC=0}
  \newline \textbf{Skill-RAG:} diagnosis \texttt{query\_misaligned}.} \\
\midrule
Round 2
& \cellcolor{red!8}%
  Query: \textit{``dancing plague Strasbourg Frau Troffea 400 people''}
  \newline \textit{Evidence: causes debated (ergotism, psychogenic illness); no confirmed disease.}
  \newline Answer: \textit{``Mass psychogenic illness''} \textcolor{red}{ACC=0}
& \cellcolor{green!8}%
  \textbf{Skill:} \texttt{query\_misaligned}
  \newline Query: \textit{``1518 victims physical exhaustion consequences''}
  \newline \textit{Evidence: victims collapsed from exhaustion; deaths confirmed.}
  \newline Answer: \textit{``exhaustion''} \textcolor{blue}{ACC=1 \checkmark}
  \newline Prober: \textbf{stop.} \\
\midrule
Round 3
& \cellcolor{red!8}%
  Query: \textit{``ergotism psychogenic illness mass hysteria 1518''}
  \newline \textit{Evidence: exact cause remains debated.}
  \newline Answer: \textit{``Ergotism''} \textcolor{red}{ACC=0 \textbf{WRONG}}
& \cellcolor{green!8}%
  Already stopped at Round 2.
  \newline \textbf{Final: exhaustion} \textcolor{blue}{\checkmark} \\
\bottomrule
\end{tabular}
\caption{Skill routing prevents query drift and recovers
the correct answer where Probing-RAG fails.}
\label{tab:case_study}
\end{table*}

\subsection{Hyperparameters}
\label{app:hyperparameters}
Table~\ref{tab:hyperparameters} summarizes the key parameters of our experimental setup, with additional implementation details available in the handbook provided in the repository.

\begin{table}
\centering
\setlength{\tabcolsep}{3.2pt}
\renewcommand{\arraystretch}{1.1}
\scriptsize
\resizebox{0.45\columnwidth}{!}{
\begin{tabular}{ll}
\toprule
Name & Value \\
\midrule
Learning rate       & $1{\times}10^{-3}$ \\
Batch size          & 12 \\
Epochs              & 2 \\
Dropout             & 0.1 \\
Activation function & SiLU \\
Normalization       & LayerNorm \\
Optimizer           & AdamW \\
Scheduler           & ExponentialLR \\
Gamma               & 0.995 \\
GPUs                & H100 $\times$ 1 \\
\bottomrule
\end{tabular}
}
\caption{Hyperparameters used for training the probe.}
\label{tab:hyperparameters}
\end{table}

\bibliography{tmlr}
\bibliographystyle{tmlr}
\end{document}